\documentclass[11pt,a4paper]{article}
\usepackage[hyperref]{acl2019}
\usepackage{times}
\usepackage{latexsym}

\usepackage{pifont}
\usepackage[normalem]{ulem}
\usepackage{microtype}
\usepackage{enumitem}
\usepackage{booktabs}
\usepackage{multirow}
\usepackage{graphicx}
\usepackage{array}
\usepackage{pgfplots}
\usepackage{subfigure}
\usepackage{amsmath}
\usepackage{algorithm}
\usepackage{url}
\usepackage{makecell}
\usepackage{subfigure}
\usepackage{amsmath}
\usepackage{amsfonts}
\usepackage{amsthm,amssymb,amsopn,bm}
\usepackage[noend]{algpseudocode}

\DeclareMathOperator*{\argmax}{arg\,max}

\makeatletter
\newcommand\footnoteref[1]{\protected@xdef\@thefnmark{\ref{#1}}\@footnotemark}
\makeatother

\aclfinalcopy 
\def\aclpaperid{1664} 

\newcommand*\samethanks[1][\value{footnote}]{\footnotemark[#1]}

\newif\ifcomments
\commentstrue 
\ifcomments
    \providecommand{\sewon}[1]{{\protect\color{blue}{[Sewon: #1]}}}
    \providecommand{\eric}[1]{{\protect\color{magenta}{[Eric: #1]}}}
    \providecommand{\matt}[1]{{\protect\color{purple}{\bf [Matt: #1]}}}
    \providecommand{\sameer}[1]{{\protect\color{purple!50!orange}{[Sameer: #1]}}}
    \providecommand{\hanna}[1]{{\protect\color{purple}{\bf [Hanna: #1]}}}
    \providecommand{\luke}[1]{{\protect\color{blue}{\bf [Luke: #1]}}}
\else
    \providecommand{\sewon}[1]{}
    \providecommand{\eric}[1]{}
    \providecommand{\matt}[1]{}
    \providecommand{\sameer}[1]{}
    \providecommand{\hanna}[1]{}
    \providecommand{\luke}[1]{}
\fi
\title{Compositional Questions Do Not Necessitate Multi-hop Reasoning}

\author{\makecell{Sewon Min\thanks{~~Equal Contribution.}$~~^1$, Eric Wallace\samethanks$~~^2$, Sameer Singh$^3$, \\ Matt Gardner$^2$, Hannaneh Hajishirzi$^{1,2}$, Luke Zettlemoyer$^{1}$}\\
$^1$University of Washington\\
$^2$Allen Institute for Artificial Intelligence\\
$^3$University of California, Irvine\\
{\tt sewon@cs.washington.edu, ericw@allenai.org}}

\date{}

\begin{document}

\newcommand{\singlebert}{single-paragraph \textsc{BERT}}
\newcommand{\singlebertcap}{Single-paragraph \textsc{BERT}}
\newcommand{\bert}{\textsc{BERT}}
\newcommand{\bertbase}{\textsc{BERT-base}}
\newcommand{\hotpot}{\textsc{HotpotQA}}
\newcommand{\wikihop}{\textsc{WikiHop}}
\newcommand{\wq}{\textsc{WebQuestions}}
\newcommand{\cwq}{\textsc{ComplexWebQuestions}}
\newcommand{\rc}{RC}
\newcommand{\tfidf}{\textsc{TF-IDF}}
\providecommand{\example}[1]{{\textsl{#1}}}
\newcommand{\cata}{27}
\newcommand{\catb}{35}
\newcommand{\catd}{26}
\newcommand{\cate}{8}
\newcommand{\catp}{4}

\maketitle
\begin{abstract}
Multi-hop reading comprehension (RC) questions are challenging because they require reading and reasoning over multiple paragraphs. We argue that it can be difficult to construct large multi-hop RC datasets.
For example, even highly compositional questions can be answered with a single hop if they target specific entity types, or the facts needed to answer them are redundant.
Our analysis is centered on \hotpot{}, where we show that single-hop reasoning can solve much more of the dataset than previously thought. We introduce a single-hop BERT-based RC model that achieves 67 F1---comparable to state-of-the-art multi-hop models. We also design an evaluation setting where humans are not shown all of the necessary paragraphs for the intended multi-hop reasoning but can still answer over 80\% of questions. Together with detailed error analysis, these results suggest there should be an increasing focus on the role of evidence in multi-hop reasoning and possibly even a shift towards information retrieval style evaluations with large and diverse evidence collections. 
\end{abstract}

\section{Introduction}\label{sec:intro}
Multi-hop reading comprehension (RC) requires reading and aggregating information over multiple pieces of textual evidence~\citep{wikihop,hotpot,talmor2018complex}. In this work, we argue that it can be difficult to construct large multi-hop RC datasets. 
This is because multi-hop reasoning is a characteristic of both the question and the provided evidence; even highly compositional questions can be answered with a single hop if they target specific entity types, or the facts needed to answer them are redundant. For example, the question in Figure~\ref{fig:example} is compositional: a plausible solution is to find ``What animal's habitat was the R\'eserve Naturelle Lomako Yokokala established to protect?'', and then answer ``What is the former name of that animal?''. However, when considering the evidence paragraphs, the question is solvable in a single hop by finding the only paragraph that describes an animal. 

\setlength{\textfloatsep}{0.5cm} 
 \begin{figure}[t]
     \centering
     \small
     \begin{tabular}{p{7.2cm}}
         \toprule     
         \textbf{Question:} What is the former name of the animal whose habitat the R\'eserve Naturelle Lomako Yokokala was established to protect? \\
         \textbf{Paragraph 5:}~The Lomako Forest Reserve is found in Democratic Republic of the Congo. It was established in 1991 especially to protect the habitat of the Bonobo apes.\\
         \textbf{Paragraph 1:}~The bonobo (``Pan paniscus''), formerly called the \textbf{\underline{pygmy chimpanzee}} and less often, the dwarf or gracile chimpanzee, is an endangered great ape and one of the two species making up the genus ``Pan''.   \\
         \bottomrule              
     \end{tabular}
     \vskip -0.7em
     \caption{A \hotpot{} example designed to require reasoning across two paragraphs. Eight spurious additional paragraphs (not shown) are provided to increase the task difficulty. However, since only one of the ten paragraphs is about an animal, one can immediately locate the answer in \emph{Paragraph 1} using one hop. The full example is provided in Appendix~\ref{app:full-example}.}
     \label{fig:example}
 \end{figure}

Our analysis is centered on \hotpot{}~\cite{hotpot}, a dataset of mostly compositional questions. In its \rc{} setting, each question is paired with two gold paragraphs, which should be needed to answer the question, and eight distractor paragraphs, which provide irrelevant evidence or incorrect answers. We show that single-hop reasoning can solve much more of this dataset than previously thought. First, we design a single-hop QA model based on~\bert~\citep{bert}, which, despite having no ability to reason across paragraphs, achieves performance competitive with the state of the art.
Next, we present an evaluation demonstrating that humans can solve over 80\% of questions when we withhold one of the gold paragraphs.

To better understand these results, we present a detailed analysis of why single-hop reasoning works so well. We show that questions include redundant facts which can be ignored when computing the answer, and that the fine-grained entity types present in the provided paragraphs in the \rc{} setting often provide a strong signal for answering the question, e.g., there is only one animal in the given paragraphs in Figure~\ref{fig:example}, allowing one to immediately locate the answer using one hop.

This analysis shows that more carefully chosen distractor paragraphs would induce questions that require multi-hop reasoning. We thus explore an alternative method for collecting distractors based on adversarial paragraph selection. Although this appears to mitigate the problem, a single-hop model re-trained on these distractors can recover most of the original single-hop accuracy, indicating that these distractors are still insufficient. Another method is to consider very large distractor sets such as all of Wikipedia or the entire Web, as done in open-domain \hotpot{} and ComplexWebQuestions~\cite{talmor2018complex}. However, this introduces additional computational challenges and/or the need for retrieval systems. Finding a small set of distractors that induce multi-hop reasoning remains an open challenge that is worthy of follow up work. 
\section{Related Work}\label{sec:related}
Large-scale \rc{} datasets~\citep{cnndailymail,rajpurkar2016squad,triviaqa} have enabled rapid advances in neural QA models~\citep{bidaf,dcn+,fast-and-accurate,bert}. To foster research on reasoning across \emph{multiple} pieces of text, multi-hop QA has been introduced~\cite{narrativeqa,talmor2018complex,hotpot}. These datasets contain compositional or ``complex'' questions. We demonstrate that these questions do not necessitate multi-hop reasoning.

Existing multi-hop QA datasets are constructed using knowledge bases, e.g., \wikihop{}~\cite{wikihop} and \cwq{}~\cite{talmor2018complex}, or using crowd workers, e.g., ~\hotpot{}~\cite{hotpot}. \wikihop{} questions are posed as triples of a relation and a head entity, and the task is to determine the tail entity of the relationship. \cwq{} consists of open-domain compositional questions, which are constructed by increasing the complexity of SPARQL queries from~\wq{}~\citep{webquestions}.
We focus on \hotpot{}, which consists of multi-hop questions written to require reasoning over two paragraphs from Wikipedia. 

Parallel research from \citet{chen2019understanding} presents similar findings on \hotpot{}. Our work differs because we conduct human analysis to understand why questions are solvable using single-hop reasoning. Moreover, we show that selecting distractor paragraphs is difficult using current retrieval methods.
\section{Single-paragraph QA}\label{sec:singlehop}
This section shows the performance of a single-hop model on \hotpot{}.

\setlength{\textfloatsep}{0.5cm} 
\begin{figure}[tb]
\centering
\resizebox{\columnwidth}{!}{
\includegraphics[width=\textwidth]{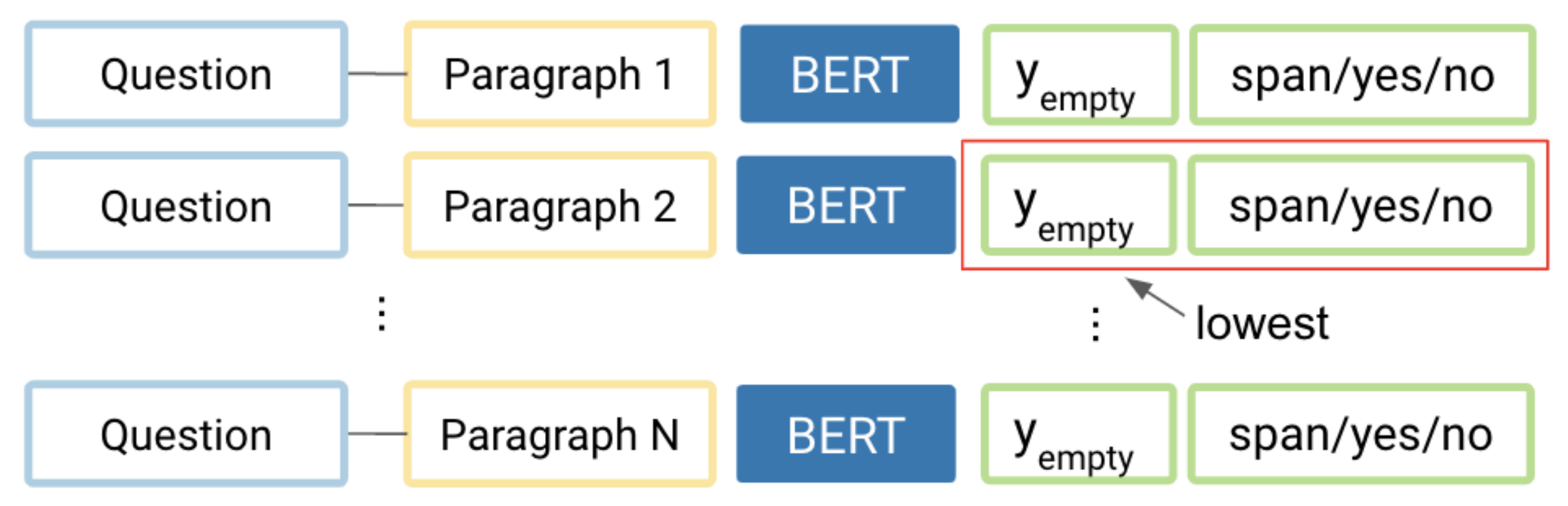}
}
\caption{Our model, \singlebert{}, reads and scores each paragraph independently. The answer from the paragraph with the lowest $y_\mathrm{empty}$ score is chosen as the final answer.}
\label{fig:singlebert}
\end{figure}

\subsection{Model Description}\label{subsec:singlehop}

Our model, \singlebert{}, scores and answers each paragraph independently (Figure~\ref{fig:singlebert}). We then select the answer from the paragraph with the best score, similar to \citet{simple-and-effective}.\footnote{Full details in Appendix~\ref{app:full-model}. Code available at \url{https://github.com/shmsw25/single-hop-rc}.}

The model receives a question $Q = [q_1, .., q_m]$ and a single paragraph $P = [p_1, ..., p_n]$ as input. Following \citet{bert}, $S = [q_1, ..., q_m, {\tt [SEP]}, p_1, ..., p_n]$, where ${\tt [SEP]}$ is a special token, is fed into~\bert:\begin{equation*}S' = \bert (S) \in \mathbb{R}^{h \times (m+n+1)}, \end{equation*}
where $h$ is the hidden dimension of~\bert. Next, a classifier uses max-pooling and learned parameters $W_1 \in \mathbb{R}^{h \times 4}$ to generate four scalars:\begin{equation*} [y_\mathrm{span}; y_\mathrm{yes}; y_\mathrm{no}; y_\mathrm{empty}]  = W_1 \mathrm{maxpool}(S'),
\end{equation*}

\noindent
where $y_\mathrm{span}, y_\mathrm{yes}, y_\mathrm{no}$ and $ y_\mathrm{empty}$ indicate the answer is either a span, \texttt{yes}, \texttt{no}, or no answer. An extractive paragraph span, $\mathrm{span}$, is obtained separately following~\citet{bert}. The final model outputs are a scalar value $y_\mathrm{empty}$ and a text of either $\mathrm{span}$, \texttt{yes} or \texttt{no}, based on which of $y_\mathrm{span}, y_\mathrm{yes}, y_\mathrm{no}$ has the largest value.

For a particular \hotpot{} example, we run \singlebert{} on each paragraph in parallel and select the answer from the paragraph with the smallest $y_\mathrm{empty}$.

\subsection{Model Results}\label{subsec:singlehop-exp}

\setlength{\textfloatsep}{0.5cm} 
\setlength{\tabcolsep}{4pt}
\begin{table}
    \footnotesize
    \centering
    \begin{tabular}{lcc}
        \toprule
        \bf Model & \bf Distractor F1 & \bf Open F1 \\
        \midrule    
        \singlebertcap * & 67.08 & 38.40 \\
        \addlinespace
        BiDAF* & 58.28 & 34.36 \\
        BiDAF & 58.99 & 32.89 \\
        GRN & 66.71 & 36.48 \\
        QFE & 68.06 & 38.06 \\
        DFGN + \bert{} & 68.49	& - \\
        MultiQA & - & 40.23 \\
        DecompRC & 69.63 & 40.65 \\
        \bert{} Plus & 69.76 & -  \\
        Cognitive Graph & - & 48.87 \\
        \bottomrule
    \end{tabular}
    \caption{F1 scores on \hotpot{}. * indicates the result is on the validation set; the other results are on the hidden test set shown in the official leaderboard.}
    \label{table:result-hotpot}
\end{table}

\hotpot{} has two settings: a distractor setting and an open-domain setting.~\smallskip

\paragraph{Distractor Setting} The \hotpot{} distractor setting pairs the two paragraphs the question was written for (\emph{gold paragraphs}) with eight spurious paragraphs selected using \tfidf{} similarity with the question (\emph{distractors}). Our \singlebert{} model achieves 67.08 F1, comparable to the state-of-the-art (Table~\ref{table:result-hotpot}).\footnote{Results as of March 4th, 2019.} This indicates the majority of \hotpot{} questions are answerable in the distractor setting using a single-hop model.~\smallskip 

\paragraph{Open-domain Setting} The \hotpot{} open-domain setting (\emph{Fullwiki}) does not provide a set of paragraphs---all of Wikipedia is considered. We follow~\citet{squad-open} and retrieve paragraphs using bigram \tfidf{} similarity with the question.

We use the \singlebert{} model trained in the distractor setting. We also fine-tune the model using incorrect paragraphs selected by the retrieval system. In particular, we retrieve 30 paragraphs and select the eight paragraphs with the lowest $y_\mathrm{empty}$ scores predicted by the trained model. 
\singlebertcap{} achieves 38.06 F1 in the open-domain setting (Table~\ref{table:result-hotpot}). This shows that the open-domain setting is challenging for our single-hop model and is worthy of future study.
\section{Compositional Questions Are Not Always Multi-hop}\label{sec:analysis}
This section provides a human analysis of \hotpot{} to understand what phenomena enable single-hop answer solutions.
\hotpot{} contains two question types, \emph{Bridge} and \emph{Comparison}, which we evaluate separately.

\subsection{Categorizing Bridge Questions}\label{subsec:bridge}

Bridge questions consist of two paragraphs linked by an entity~\cite{hotpot}, e.g., Figure~\ref{fig:example}. We first investigate single-hop human performance on \hotpot{} bridge questions using a human study consisting of NLP graduate students. Humans see the paragraph that contains the answer span and the eight distractor paragraphs, but do not see the other gold paragraph. As a baseline, we show a different set of people the same questions in their standard ten paragraph form.

On a sample of 200 bridge questions from the validation set, human accuracy shows marginal degradation when using only one hop: humans obtain \textbf{87.37 F1} using all ten paragraphs and \textbf{82.06 F1} when using only nine (where they only see a single gold paragraph). This indicates humans, just like models, are capable of solving bridge questions using only one hop.

Next, we manually categorize what enables single-hop answers for 100 bridge validation examples (taking into account the distractor paragraphs), and place questions into four categories (Table~\ref{tab:labeling}).~\smallskip 
\setlength{\textfloatsep}{0.5cm} 
\begin{table*}
    \footnotesize
    \centering
    \begin{tabular}{clc}
        \toprule
       {\bf Type} & {\bf Question} & {\bf \%} \\
        \midrule
        {Multi-hop} & Ralph Hefferline was a psychology professor at a university that is located in what city? & {\cata} \\
        \addlinespace
        \multirow{2}{*}{Weak distractors} & What government position was held by the woman who portrayed Corliss Archer in & \multirow{2}{*}{\catb} \\
        & the film Kiss and Tell? & \\
        \addlinespace
        \multirow{2}{*}{Redundant evidence}  & Kaiser Ventures corporation was founded by an American industrialist who became 
        & \multirow{2}{*}{\catd} \\
        & known as the father of modern American shipbuilding?\\
        \addlinespace
        Non-compositional 1-hop & {When was Poison's album `Shut Up, Make Love' released?} & {\cate}\\
        \bottomrule
    \end{tabular}
        \caption{We categorize bridge questions while taking the paragraphs into account. We exclude \catp\% of questions that we found to have incorrect or ambiguous answer annotations. See Section~\ref{subsec:bridge} for details on question types.}
        \label{tab:labeling}
\end{table*}

\begin{table*}
\footnotesize
\centering
\begin{tabular}{clcc}
    \toprule
    {\bf Type} & {\bf Question} & {\bf \%} & \bf F1 \\
    \midrule
    {Multi-hop}& Who was born first, Arthur Conan Doyle or Penelope Lively?  & {45} & {54.46}\\
    \addlinespace
    Context-dependent& Are Hot Rod and the Memory of Our People both magazines?& {36} & {56.16}\\
    \addlinespace
    {Single-hop} & Which writer was from England, Henry Roth or Robert Erskine Childers?& {17} & {70.54}\\
    \bottomrule
\end{tabular}
    \caption{
     We automatically categorize comparison questions using rules (2\% cannot be automatically categorized).
     \singlebertcap~achieves near chance accuracy on multi-hop questions but exploits single-hop ones.
    }\label{tab:comparison}
\end{table*}

\paragraph{Multi-hop} $\cata\%$ of questions require multi-hop reasoning. The first example of Table~\ref{tab:labeling} requires locating the university where ``Ralph Hefferline'' was a psychology professor, and multiple universities are provided as distractors. Therefore, the answer cannot be determined in one hop.\footnote{It is possible that a single-hop model can do well by randomly guessing between two or three well-typed options, but we do not evaluate that strategy here.} ~\smallskip 

\paragraph{Weak Distractors} $\catb\%$ of questions allow single-hop answers in the distractor setting, mostly by entity type matching. Consider the question in the second row of Table~\ref{tab:labeling}: in the ten provided paragraphs, only one actress has a government position. Thus, the question is answerable without considering the film ``Kiss and Tell.'' These examples may become multi-hop in the open-domain setting, e.g., there are numerous actresses with a government position on Wikipedia.~\smallskip 

To further investigate entity type matching, we reduce the question to the first five tokens starting from the wh-word, following \citet{sugawara2018easier}. Although most of these reduced questions appear void of critical information, the F1 score of~\singlebert~only degrades about 15 F1 from 67.08 to 52.13.

\paragraph{Redundant Evidence} $\catd\%$ of questions are compositional but are solvable using only part of the question. For instance, in the third example of Table~\ref{tab:labeling} there is only a single founder of ``Kaiser Ventures.'' Thus, one can ignore the condition on ``American industrialist'' and ``father of modern American shipbuilding.'' This category differs from the weak distractors category because its questions are single-hop regardless of the distractors.~\smallskip 

\paragraph{Non-compositional Single-hop} $\cate\%$ of questions are non-compositional and single-hop. In the last example of Table~\ref{tab:labeling}, one sentence contains all of the information needed to answer correctly.

\subsection{Categorizing Comparison Questions}\label{subsec:comparison}

Comparison questions require quantitative or logical comparisons between two quantities or events. We create rules (Appendix~\ref{app:comparison}) to group comparison questions into three categories: questions which require multi-hop reasoning (\emph{multi-hop}), may require multi-hop reasoning (\emph{context-dependent}), and require single-hop reasoning (\emph{single-hop}). 

Many comparison questions are multi-hop or context-dependent multi-hop, and \singlebert{} achieves near chance accuracy on these types of questions (Table~\ref{tab:comparison}).\footnote{Comparison questions test mainly binary relationships.} This shows that most comparison questions are not solvable by our single-hop model.
\section{Can We Find Better Distractors?}\label{sec:correcting}
In Section~\ref{subsec:bridge}, we identify that \catb{}\% of bridge examples are solvable using single-hop reasoning due to weak distractor paragraphs.
Here, we attempt to automatically correct these examples by choosing new distractor paragraphs which are likely to trick our single-paragraph model.

\begin{table}
\footnotesize
\centering
\begin{tabular}{lcc}
        \toprule
        \multirow{2}{*}{Evaluation Data} & \multicolumn{2}{c}{Training Data} \\
        \cmidrule{2-3}
        & Original & Adversarial \\
        \midrule
        Original & 67.08 & 59.12\\
        Adversarial & 46.84 & 60.10\\
        + Type & 40.73 & 58.42\\        
        \bottomrule
    \end{tabular}
    \caption{We train on \hotpot{} using standard distractors (\emph{Original}) or using adversarial distractors (\emph{Adversarial}). The model is then tested on the original distractors, adversarial distractors, or adversarial distractors with filtering by entity type (\emph{+ Type}).} \label{table:distractors}
\end{table}

\paragraph{Adversarial Distractors} We select the top-50 first paragraphs of Wikipedia pages using \tfidf{} similarity with the question, following the original \hotpot{} setup. 
Next, we use \singlebert{} to adversarially select the eight distractor paragraphs from these 50 candidates. In particular, we feed each paragraph to the model and select the paragraphs with the lowest $y_\mathrm{empty}$ score (i.e., the paragraphs that the model thinks contain the answer). These paragraphs are dissimilar to the original distractors---there is a 9.82\% overlap.

We report the F1 score of \singlebert{} on these new distractors in Table~\ref{table:distractors}: the accuracy declines from 67.08 F1 to 46.84 F1. However, when the same procedure is done on the training set and the model is re-trained, the accuracy increases to 60.10 F1 on the adversarial distractors. 

\paragraph{Type Distractors} We also experiment with filtering the initial list of 50 paragraph to ones whose entity type (e.g., person) matches that of the gold paragraphs. This can help to eliminate the entity type bias described in Section~\ref{subsec:bridge}. As shown in Table~\ref{table:distractors}, the original model's accuracy degrades significantly (drops to 40.73 F1). However, similar to the previous setup, the model trained on the adversarially selected distractors can recover most of its original accuracy (increases to 58.42 F1).

These results show that \singlebert{} can struggle when the distribution of the distractors changes (e.g., using adversarial selection rather than only \tfidf{}). Moreover, the model can somewhat recover its original accuracy when re-trained on distractors from the new distribution. 
\section{Conclusions}\label{sec:discussion}
In summary, we demonstrate that question compositionality is not a sufficient condition for multi-hop reasoning. Instead, future datasets must carefully consider what evidence they provide in order to ensure multi-hop reasoning is required. There are at least two different ways to achieve this.

\paragraph{Open-domain Questions} Our single-hop model struggles in the open-domain setting. We largely attribute this to the insufficiencies of standard \tfidf{} retrieval for multi-hop questions.
For example, we fail to retrieve the paragraph about ``Bonobo apes'' in Figure~\ref{fig:example}, because the question does not contain terms about ``Bonobo apes.''
Table~\ref{tab:open_domain} shows that the model achieves 39.12 F1 given 500 retrieved paragraphs, but achieves 53.12 F1 when additional two gold paragraphs are given, demonstrating the significant effect of failure to retrieve gold paragraphs.
In this context, we suggest that future work can explore better retrieval methods for multi-hop questions. 

\paragraph{Retrieving Strong Distractors} Another way to ensure multi-hop reasoning is to select strong distractor paragraphs.
For example, we found $\catb\%$ of bridge questions are currently single-hop but may become multi-hop when combined with stronger distractors (Section~\ref{subsec:bridge}).
However, as we demonstrate in Section~\ref{sec:correcting}, selecting strong distractors for \rc{} questions is non-trivial. We suspect this is also due to the insufficiencies of standard \tfidf{} retrieval for multi-hop questions.
In particular, Table~\ref{tab:open_domain} shows that \singlebert{} achieves 53.12 F1 even when using 500 distractors (rather than eight), indicating that 500 distractors are still insufficient.
In this end, future multi-hop RC datasets can develop improved methods for distractor collection.

\begin{table}
\footnotesize
\centering
\begin{tabular}{lcc}
        \toprule
        \bf Setting & \bf F1 \\
        \midrule    
        Distractor & 67.08\\
        \addlinespace    
        Open-domain 10 Paragraphs & 38.40 \\
        Open-domain 500 Paragraphs & 39.12 \\ 
        \hspace{0.3cm} + Gold Paragraph & 53.12 \\
        \bottomrule
    \end{tabular}
    \caption{The accuracy of \singlebert{} in different open-domain retrieval settings. \tfidf{} often fails to retrieve the gold paragraphs even when using 500 candidates.} \label{tab:open_domain}
\end{table}
\section*{Acknowledgements}
This research was supported by ONR (N00014-18-1-2826, N00014-17-S-B001), NSF (IIS-1616112, IIS-1252835, IIS-1562364), ARO (W911NF-16-1-0121), an Allen Distinguished Investigator Award, Samsung GRO and gifts from Allen Institute for AI, Google, and Amazon. 

The authors would like to thank Shi Feng, Nikhil Kandpal, Victor Zhong, the members of AllenNLP and UW NLP,
and the anonymous reviewers for their valuable feedback.

\bibliography{journal-abbrv,bib}
\bibliographystyle{acl_natbib}

\clearpage
\appendix
\section{Example Distractor Question}\label{app:full-example}

We present the full example from Figure~\ref{fig:example} below. Paragraphs 1 and 5 are the two gold paragraphs.

\paragraph{Question} What is the former name of the animal whose habitat the R\'eserve Naturelle Lomako Yokokala was established to protect?

\paragraph{Answer} pygmy chimpanzee

\paragraph{(Gold Paragraph) Paragraph 1} The bonobo (or ; ``Pan paniscus''), formerly called the \textbf{\underline{pygmy chimpanzee}} and less often, the dwarf or gracile chimpanzee, is an endangered great ape and one of the two species making up the genus ``Pan''; the other is ``Pan troglodytes'', or the common chimpanzee.  Although the name ``chimpanzee'' is sometimes used to refer to both species together, it is usually understood as referring to the common chimpanzee, whereas ``Pan paniscus'' is usually referred to as the bonobo.

\paragraph{Paragraph 2} The Carri\'ere des Nerviens Regional Nature Reserve (in French ``R\'eserve naturelle r\'egionale de la carri\'ere des Nerviens'') is a protected area in the Nord-Pas-de-Calais region of northern France.  It was established on 25 May 2009 to protect a site containing rare plants and covers just over 3 ha.  It is located in the municipalities of Bavay and Saint-Waast in the Nord department.

\paragraph{Paragraph 3} C\'ereste (Occitan: ``Ceir\'esta'') is a commune in the Alpes-de-Haute-Provence department in southeastern France.  It is known for its rich fossil beds in fine layers of ``Calcaire de Campagne Calavon'' limestone, which are now protected by the Parc naturel r\'egional du Luberon and the R\'eserve naturelle g\'eologique du Luberon.

\paragraph{Paragraph 4} The Grand Cote National Wildlife Refuge (French: ``R\'eserve Naturelle Faunique Nationale du Grand- Cote'') was established in 1989 as part of the North American Waterfowl Management Plan.  It is a 6000 acre reserve located in Avoyelles Parish, near Marksville, Louisiana, in the United States.

\paragraph{(Gold Paragraph) Paragraph 5} The Lomako Forest Reserve is found in Democratic Republic of the Congo.  It was established in 1991 especially to protect the habitat of the Bonobo apes.  This site covers 3,601.88 km$^2$.

\paragraph{Paragraph 6} Guadeloupe National Park (French: ``Parc national de la Guadeloupe'') is a national park in Guadeloupe, an overseas department of France located in the Leeward Islands of the eastern Caribbean region.  The Grand Cul-de-Sac Marin Nature Reserve (French: ``R\'eserve Naturelle du Grand Cul-de-Sac Marin'') is a marine protected area adjacent to the park and administered in conjunction with it.  Together, these protected areas comprise the Guadeloupe Archipelago (French: ``l'Archipel de la Guadeloupe'') biosphere reserve.

\paragraph{Paragraph 7} La D\'esirade National Nature Reserve (French: ``R\'eserve naturelle nationale de La D\'esirade'') is a reserve in D\'esirade Island in Guadeloupe.  Established under the Ministerial Decree No. 2011-853 of 19 July 2011 for its special geological features it has an area of 62 ha.  The reserve represents the geological heritage of the Caribbean tectonic plate, with a wide spectrum of rock formations, the outcrops of volcanic activity being remnants of the sea level oscillations. It is one of thirty three geosites of Guadeloupe.

\paragraph{Paragraph 8} La Tortue ou l'Ecalle or Ile Tortue is a small rocky islet off the northeastern coast of Saint Barth\'elemy in the Caribbean.  Its highest point is 35 m above sea level.  Referencing tortoises, it forms part of the R\'eserve naturelle nationale de Saint-Barth\'elemy with several of the other northern islets of St Barts.

\paragraph{Paragraph 9} Nature Reserve of Saint Bartholomew (R\'eserve Naturelle de Saint-Barth\'elemy) is a nature reserve of Saint Barth\'elemy (RNN 132), French West Indies, an overseas collectivity of France.

\paragraph{Paragraph 10} Ile Fourchue, also known as Ile Fourche is an island between Saint-Barth\'elemy and Saint Martin, belonging to the Collectivity of Saint Barth\'elemy.  The island is privately owned.  The only inhabitants are some goats.  The highest point is 103 meter above sea level.  It is situated within R\'eserve naturelle nationale de Saint-Barth\'elemy.

\section{Full Model Details}\label{app:full-model}

\begin{figure*}[!ht]
\centering
\resizebox{1.6\columnwidth}{!}{
\includegraphics[width=\textwidth]{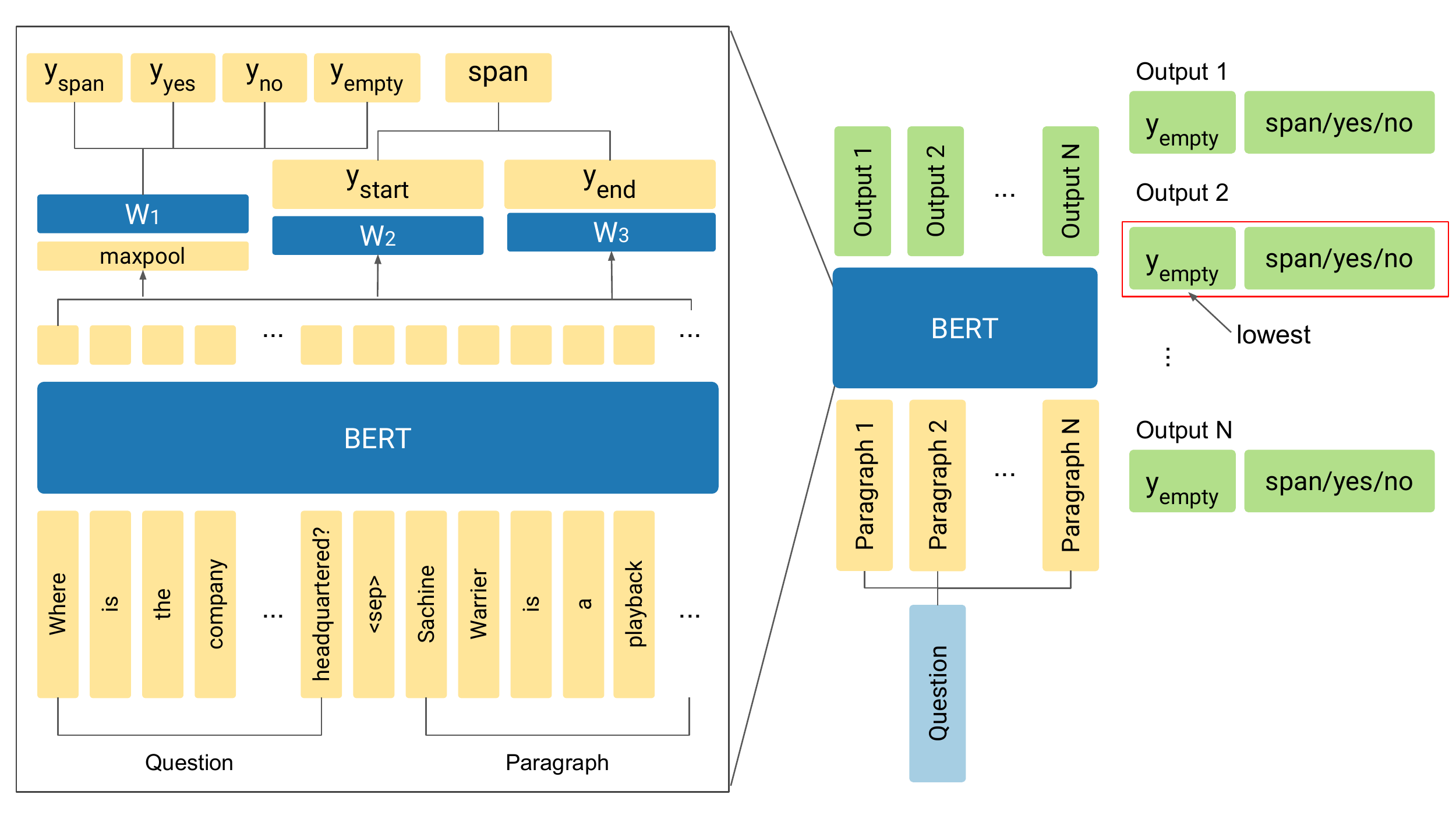}
}
\caption{Single-paragraph \textsc{BERT} reads and scores each paragraph independently. The answer from the paragraph with the lowest $y^\mathrm{empty}$ score is chosen as the final answer.}

\label{fig:singlebertfull}
\end{figure*}

Single-paragraph BERT is a pipeline which first retrieves a single paragraph using a classifier and then selects the associated answer. Formally, the model receives a question $Q = [q_1, .., q_m]$ and a single paragraph $P = [p_1, ..., p_n]$ as input. The question and paragraph are merged into a single sequence, $S = [q_1, ..., q_m, {\tt [SEP]}, p_1, ..., p_n]$, where ${\tt [SEP]}$ is a special token indicating the boundary. The sequence is fed into~\bertbase:
\begin{equation}\nonumber
    S' = \bert(S) \in \mathbb{R}^{h \times (m+n+1)},
\end{equation}

where $h$ is the hidden dimension of~\bert. Next, a classifier uses max-pooling and learned parameters $W_1 \in \mathbb{R}^{h \times 4}$ to generate four scalars:\begin{equation}\nonumber
    [y_\mathrm{span}; y_\mathrm{yes}; y_\mathrm{no}; y_\mathrm{empty}]  = W_1 \mathrm{maxpool}(S'),
\end{equation}

where $y_\mathrm{span}, y_\mathrm{yes}, y_\mathrm{no}$ and $ y_\mathrm{empty}$ indicate the answer is either a span, \texttt{yes}, \texttt{no}, or no answer. 

A candidate answer span is then computed separately from the classifier. We define\begin{align*}\nonumber
    p_\mathrm{start} = \mathrm{Softmax} (W_2 S') \\ 
    p_\mathrm{end} = \mathrm{Softmax} (W_3 S'),
\end{align*}
where $W_2, W_3 \in \mathbb{R}^{h}$ are learned parameters.
Then, $y_\mathrm{start}$ and $y_\mathrm{end}$ are obtained:\begin{align*}\nonumber
    y_\mathrm{start}, y_\mathrm{end} = \argmax_{i \leq j} p_\mathrm{start}^i p_\mathrm{end}^j
\end{align*}
where $p_\mathrm{start}^i$ and $p_\mathrm{end}^j$ indicate the $i$-th element of $p_\mathrm{start}$ and $j$-th element of $p_\mathrm{end}$, respectively.

We now have four scalar values $y_\mathrm{span}$, $y_\mathrm{yes}$, $y_\mathrm{no}$, and $y_\mathrm{empty}$ and a span from the paragraph $\mathrm{span}=[S_{y_\mathrm{start}}, \dots, S_{y_\mathrm{end}}]$.

For \hotpot{}, the input is a question and $N$ context paragraphs. We create a batch of size $N$, where each entry is a question and a single paragraph. Denote the ouput from $i$-th entry as $y^i_\mathrm{span}, y^i_\mathrm{yes}, y^i_\mathrm{no}, y^i_\mathrm{empty}$and $\mathrm{span}^i$. The final answer is selected as:\begin{eqnarray}\nonumber
    j &=& \mathrm{argmin}_i (y^i_\mathrm{empty}) \\ \nonumber
    y_\mathrm{max} &=& \mathrm{max} (y^j_\mathrm{span}, y^j_\mathrm{yes}, y^j_\mathrm{no})\\
    \mathrm{answer} &=& \begin{cases}\nonumber
        \mathrm{span}^j  & \mathrm{if} y^j_\mathrm{span} = y_\mathrm{max} \\
        \texttt{yes} & \mathrm{if} y^j_\mathrm{yes} = y_\mathrm{max} \\
        \texttt{no}  & \mathrm{if} y^j_\mathrm{no} = y_\mathrm{max}
    \end{cases}
\end{eqnarray}
\noindent During training, $y^i_\mathrm{empty}$ is set to 0 for the paragraph which contains the answer span and 1 otherwise. 

\paragraph{Implementation Details} We use PyTorch~\citep{pytorch} based on Hugging Face's implementation.\footnote{\url{https://github.com/huggingface/pytorch-pretrained-BERT}} We use Adam~\citep{adam} with learning rate $5\times10^{-5}$. We lowercase the input and set the maximum sequence length $\vert{S}\vert$ to $300$. If a sequence is longer than $300$, we split it into multiple sequences and treat them as different examples.

\section{Categorizing Comparison Questions}\label{app:comparison}

\begin{table*}[t]
    \centering
    \footnotesize
    \begin{tabular}{l} 
     \toprule
     Operation \& Example \\
     \midrule
        \textbf{Numerical Questions}\\
        Operations: {\tt Is greater / Is smaller / Which is greater / Which is smaller} \\
        Example (\texttt{Which is smaller}): Who was born first, Arthur Conan Doyle or Penelope Lively? \\
     \midrule
        \textbf{Logical Questions}\\
        Operations: {\tt And / Or / Which is true} \\
        Example (\texttt{And}): Are Hot Rod and the Memory of Our People both magazines?\\
     \midrule
        \textbf{String Questions}\\
        Operations: {\tt Is equal / Not equal / Intersection} \\
        Example (\texttt{Is equal}): Are {Cardinal Health} and {Kansas City Southern} located in the same state? \\
     \bottomrule
\end{tabular}
\caption{The question operations used for categorizing comparison questions.} 
\label{tab:operations}
\vspace{-8pt}
\end{table*}

\begin{algorithm*}
\small
\caption{Algorithm for Identifying Question Operations}\label{alg:comparison-heuristics}
\begin{algorithmic}[1]
\Procedure{Categorize}{question, entity1, entity2}
\State coordination, preconjunct $\gets f$(question, entity1, entity2) 
\State Determine if the question is {\em either} question or {\em both} question from coordination and preconjunct
\State head entity $\gets f_{head}$(question, entity1, entity2)
\If {{\em more, most, later, last, latest, longer, larger, younger, newer, taller, higher} in question}
    \If {head entity exists}
        discrete\_operation $\gets$ Which is greater
    \Else{}
        discrete\_operation $\gets$ Is greater
    \EndIf {}
\ElsIf {{\em less, earlier, earliest, first, shorter, smaller, older, closer} in question}   
    \If {head entity exists}
        discrete\_operation $\gets$ Which is smaller
    \Else{}
        discrete\_operation $\gets$ Is smaller
    \EndIf {}
\ElsIf {head entity exists}
    \State discrete\_operation $\gets$ Which is true
\ElsIf {question is not yes/no question and asks for the property in common}
    \State discrete\_operation $\gets$ Intersection
\ElsIf {question is yes/no question}
    \State Determine if question asks for logical comparison or string comparison
    \If {question asks for logical comparison}
        \If {{\em either} question}
        discrete\_operation $\gets$ Or
        \ElsIf {{\em both} question}
            discrete\_operation $\gets$ And
        \EndIf {}
    \ElsIf {question asks for string comparison}
        \If {asks for same?}
            discrete\_operation $\gets$ Is equal
        \ElsIf {asks for difference?}
            discrete\_operation $\gets$ Not equal
        \EndIf {}
    \EndIf {}
\EndIf
\State \Return discrete\_operation
\EndProcedure
\end{algorithmic}
\end{algorithm*}

This section describes how we categorize comparison questions. We first identify ten question operations that sufficiently cover comparison questions (Table~\ref{tab:operations}). Next, for each question, we extract the two entities under comparison using the Spacy\footnote{\url{https://spacy.io/}} NER tagger on the question and the two \hotpot{} supporting facts.  Using these extracted entities, we identity the suitable question operation following Algorithm~\ref{alg:comparison-heuristics}.

Based on the identified operation, questions are classified into multi-hop, context-dependent multi-hop, or single-hop. First, numerical questions are always multi-hop (e.g., first example of Table~\ref{tab:operations}). Next, the operations {\tt And, Or, Is equal,} and {\tt Not equal} are context-dependent multi-hop. For instance, in the second example of Table~\ref{tab:operations}, if ``Hot Rod'' is not a magazine, one can immediately answer \texttt{No}. Finally, the operations {\tt Which is true} and {\tt Intersection} are single-hop because they can be answered using one paragraph regardless of the context. For instance, in the third example of Table~\ref{tab:operations}, if Henry Roth's paragraph explains he is from England, one can answer Henry Roth, otherwise, the answer is Robert Erskine Childers.

\end{document}